\documentclass[conference]{IEEEtran}
\IEEEoverridecommandlockouts
\usepackage{cite}
\usepackage{amsmath,amssymb,amsfonts}
\usepackage{algorithmic}
\usepackage{graphicx}
\usepackage{textcomp}
\usepackage{xcolor}
\usepackage{pgfplots}
\pgfplotsset{compat=1.14}
\usepackage{tikz}
\usetikzlibrary{plotmarks}
\usepackage{subfiles}
\def\BibTeX{{\rm B\kern-.05em{\sc i\kern-.025em b}\kern-.08em
    T\kern-.1667em\lower.7ex\hbox{E}\kern-.125emX}}
    
\newcommand\copyrighttext{%
  \footnotesize 978-1-7281-8750-1/20/\$31.00 \textcopyright 2020 IEEE
  }
\newcommand\copyrightnotice{%
    \begin{tikzpicture}[remember picture,overlay]
    \node[anchor=south,yshift=10pt] at (current page.south) {\copyrighttext};
    \end{tikzpicture}%
}

\begin{document}

\title{Cloud Cover Nowcasting with Deep Learning}

\author{\IEEEauthorblockN{L\'ea Berthomier}
\IEEEauthorblockA{\textit{AI Lab} \\
\textit{METEO FRANCE}\\
Toulouse, France \\
lea.berthomier@meteo.fr}
\and
\IEEEauthorblockN{Bruno Pradel}
\IEEEauthorblockA{\textit{AI Lab} \\
\textit{METEO FRANCE}\\
Toulouse, France \\
bruno.pradel@meteo.fr}
\and
\IEEEauthorblockN{Lior Perez}
\IEEEauthorblockA{\textit{AI Lab} \\
\textit{METEO FRANCE}\\
Toulouse, France \\
lior.perez@meteo.fr}
}

\maketitle

\copyrightnotice

\begin{abstract}
Nowcasting is a field of meteorology which aims at forecasting weather on a short term of up to a few hours. In the meteorology landscape, this field is rather specific as it requires particular techniques, such as data extrapolation, where conventional meteorology is generally based on physical modeling. In this paper, we focus on cloud cover nowcasting, which has various application areas such as satellite shots optimisation and photovoltaic energy production forecast.

Following recent deep learning successes on multiple imagery tasks, we applied Deep Convolutionnal Networks on Meteosat satellite images for cloud cover nowcasting. We present the results of several architectures specialized in image segmentation and time series prediction. We selected the best models according to machine learning metrics as well as meteorological metrics. All selected architectures showed significant improvements over persistence and the well-known U-Net surpasses AROME physical model.
\end{abstract}


\section{Introduction}

Cloud cover nowcasting aims at forecasting the position of clouds on a short time scale of up to 3 hours. Historically, the main task of weather services is to ensure the safety of life and property. While having multiple applications, cloud cover remains less critical for safety than other phenomena such as precipitation or thunderstorms which explain its relatively lower popularity in the literature. It remains a field of interest for meteorological organisations, for observation satellite management, to optimize their image shots, and for solar panels management, to forecast their production of electricity. 
Put simply, at METEO FRANCE, a nowcasting product is a fusion of two approaches: pure data extrapolation techniques, which perform better on the very short term (up to 1 hour), and a mix between extrapolation and classical physical models which remain the best option for further forecasting. Extrapolation methods use data observed in the near past: mainly satellite and radar images. In this paper, we will focus only on the data extrapolation part and the experiments exhibited in the paper only consider the next hour and a half. In order to study the cloud cover, we will use satellite images resulting from products of EUMETSAT, the European Organisation for the Exploitation of Meteorological Satellites. 

Section \ref{background} will cover the state of the art of time series prediction applied on images, which guided our choices of neural networks architectures. In Section \ref{methodology}, we describe our methodology, metrics and choices of models. Section \ref{synthdata} presents a benchmark of several models on simple synthetic data. Finally, Sections \ref{satdata}, \ref{input1},  \ref{cloudevolution} and \ref{comparison}  present our benchmark and results on satellite images, including a study on the influence of temporal depth as an input and a comparison with two meteorological models.

\section{Background and related work} \label{background}

Cloud nowcasting is mainly computed with techniques of extrapolation of pixels or objects from wind vectors. For instance \cite{b9} uses optical flow and ground-based sky imagery for short-term solar forecasting and recent work  applies extrapolation of atmospheric motion vector  for cloud classification forecasting using images from geostationary satellite \cite{b8}. However, these methods require long computation times when applied to large areas (around 10 minutes for a full scan disk), which, added to the data reception time, often makes the first forecasts useless. In addition, these methods only rely on the observed dynamics of the clouds, and therefore cannot predict evolution in terms of size and shape of clouds, as underlined in \cite{b5}. 

Deep Learning techniques, while they were only developed in the recent years, show impressive results in the field of image processing such as image classification, object detection or image segmentation, and they require very short computation times for inference tasks. Convolutional neural networks already yielded good results on optical flow computation \cite{b4} and rainfall nowcasting \cite{b12}, \cite{b13}, \cite{b6}, \cite{b10}, \cite{b11}. Both rainfall and cloud cover nowcasting aim at forecasting the future positions of moving objects and are considering time series of images as inputs and targets of computational models. While rainfall has gained a lot of attention in the recent nowcasting literature, to the best of our knowledge cloud cover nowcasting with deep learning has only been explored in the case of predictions of satellite images \cite{b14}. Consequently all rainfall deep learning nowcasting related works will be of interest. 

Nevertheless, rainfall and cloud cover differ in that cumulative rainfall is a continuous variable computed on a pixel over a period of time, while cloud cover is a discrete class observed at a given moment on one pixel (see below for a description of the Cloud Type classification product by EUMETSAT). Consequently, cloud cover classification and forecasting is akin to a task of image segmentation applied to image time series. Thus, the output of a deep learning cloud cover nowcasting model will consist in several fields of probabilities of having a cloud on each pixel, one for each time step, where rainfall nowcasting models output only the most probable value of cumulated rainfall on each pixel. This distribution of probability is an advantage of deep learning techniques over extrapolation methods, which output binary values, as it allows to take different kind of decisions depending on the problem at hand.

Finally, cumulated rainfall is measured by doppler radar every 5 minutes while cloud cover classification results from satellite images and are collected on longer time intervals, ranging from 5 to 15 minutes depending on the satellite, even if they offer larger views. For instance, Meteosat Second Generation 4 (MSG4) captures a full scan disk of the planet each 15 minutes. This difference in time step might have consequences on the performance of the models and raises the question of how much temporal depth should be given to the model as an input.

\section{Methodology} \label{methodology}

Our goal was to forecast the position of clouds on a short time scale of 1h30' from satellite images using deep convolutional networks as our models. To this end, we used the following data and methodology.

\subsection{Data}

Our data come from the "Geostationary Nowcasting Cloud Type" classification product of EUMETSAT \cite{b7}, classifying clouds in 16 classes depending on their height and type. This classification is computed with various visible and infrared channels images shot by Meteosat Second Generation (MSG), a geostationary satellite at longitude 0 degree. The result is a 3712 by 3712 pixels image computed every 15 minutes and indicating the cloud type of each pixel. Our dataset was composed of images from 2017 and 2018.

\subsection{Experimental protocol}

Our long term objective is to forecast the class, position and height of the clouds for each pixel over the whole globe. However, for the experiments presented in this paper, we focused on forecasting only the position of the clouds, by using binary images of cloud cover as our input data : the value 1 indicates a cloud on a pixel and the value 0 indicates a cloudless pixel. Due to computation issues, we also cropped and projected the satellite images to keep only squares of size 256 by 256 pixels showing France. The final resolution of the images was 4.5 kilometers and each couple of images was spaced by 15 minutes. 

The input data of the model consisted of 4 binary images (spanning 1 hour) and our goal was to forecast the cloud cover for the next 1h30. Consequently, the output of the model consisted of 6 images of values ranging between 0 and 1 and representing the probability of having a cloud on each pixel. Figure 1 shows an example of a sequence from the dataset.

Therefore, in order to train our models, we constituted a dataset of sequences of 10 binary images, spanning 2h30. Data from 2017 and the first semester of 2018 were used as training set and the second semester of 2018 was used as validation set. 

\begin{figure}[t]
  \centering
  \includegraphics[width=0.5\textwidth]{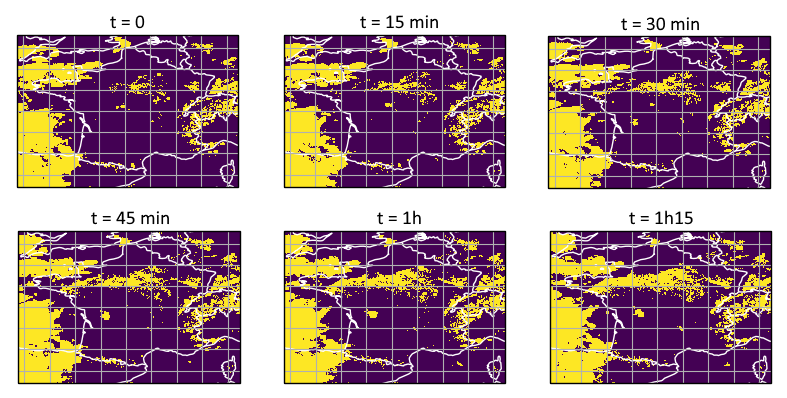}
  \caption{Example of 6 images from a sequence of the test set.}
  \label{fig:seqSat}
\end{figure}

Each of the models was trained 3 times on the dataset during 20 epochs, meaning that each sample in the dataset was seen 20 times by the network. All the following metrics were computed as means of the 3 runs.

\subsection{Metrics}

Once the models were trained with the data, we computed the mean squared error (MSE) of the models for each of the 6 output time steps. We compared these errors to the MSE of the persistence, a common meteorological baseline used in nowcasting, which consists in using the last input as prediction for all the output time steps. Indeed, in short term forecasting, the last observation is often the best prediction.  

In addition, we also computed the MSE over the binarized outputs of the model : each value of the output images was rounded to 0 if it was lower than 0.5 or 1 if it was greater than 0.5, before computing the MSE.

\begin{figure}[t]
  \centering
  \includegraphics[width=0.5\textwidth]{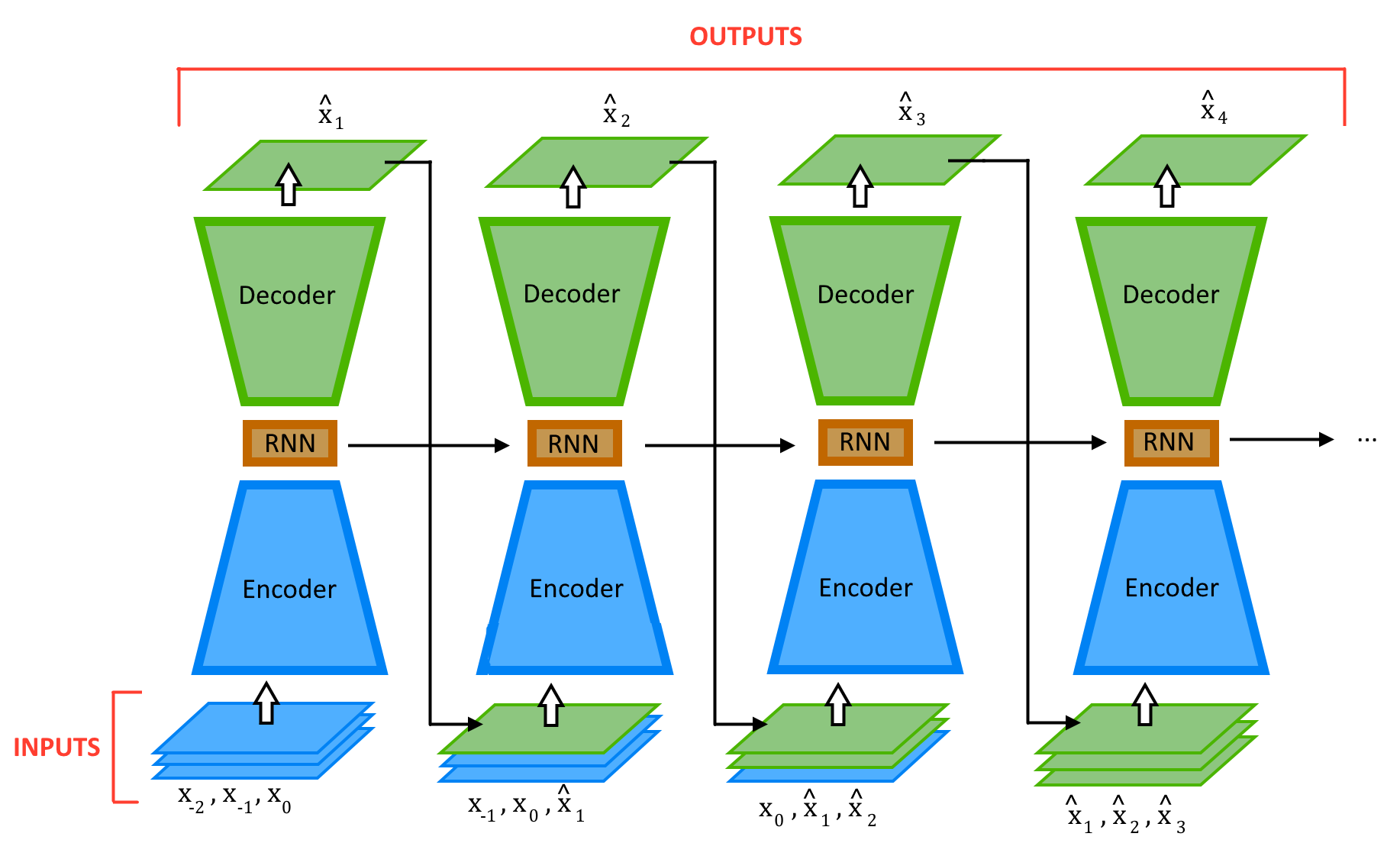}
  \caption{Diagram of the recurrent network's principle : The $x_{i}$ are the model's input and the $\hat{x_i}$ are the desired output. The output of each decoder is used as input to compute the next time step's forecast. Information between time steps is communicated through the RNN's output.}
  \label{fig:diagramRNN}
\end{figure}

\subsection{Neural networks architectures}

\begin{figure}[t]
  \centering
  \includegraphics[width=0.5\textwidth]{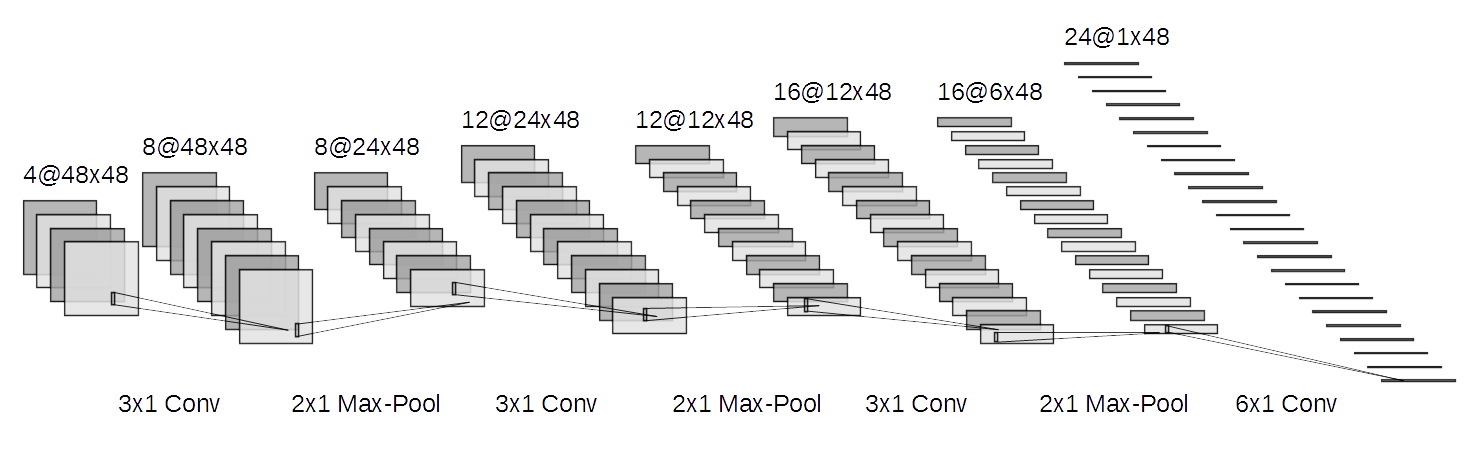}
  \caption{Reduce-LSTM - Principle of the dimension reduction.}
  \label{fig:diagramRLSTM}
\end{figure}

We built 20 different models based on the following 4 types of architecture, with several variations like the number of convolutional layers and adding residual layers or inception layers:
\begin{itemize}
    \item \textbf{Convolutional networks} with stacked convolutional layers of varying kernel size.
    
    \item \textbf{U-Net networks}, implementing the work of \cite{b2} which specializes in image segmentation, as well as variations from this architecture (adding or removing convolutional layers, removing the residuals, changing the activation function, adding fully-connected layers, ...).
    
    \item \textbf{Recurrent networks} based on the diagram on Figure \ref{fig:diagramRNN}, with variations in the architecture of the encoders and decoders.
    
    \item \textbf{LSTM networks} adapted to our data type. As we deal with time-series of 2D data, we couldn't fit them directly into a LSTM which expects 1D multi-channel data as inputs. We therefore propose the following architecture that we call \textbf{Reduce-LSTM}. We consider time as a channel and reduce the 2D input tensors of shape $X*Y*4$ to a 1D multi-channel tensors of size $Y*256$ with an encoder composed of (3,1) and (2,1) kernels for the convolutional and max pooling layers. This dimension reduction is illustrated in Figure \ref{fig:diagramRLSTM}. We then repeat this process on the input after swapping the $X$ and $Y$ axes, we stack the two resulting 1D multi-channel tensors, and use this as an input for a traditional LSTM.
\end{itemize}

\section{Benchmark on simple moving shapes} \label{synthdata}

We started our study by focusing on  synthetic images of simple moving shapes in order to check that our models could perform a simple task of time series prediction. The dataset was made of 5000 sequences of 10 images showing simple shapes like squares or circles following randomly generated rotations and linear movements. We made six datasets by varying the speed, transparency and size of the shapes, in order to test the limits of the models.

\begin{figure}[t]
\centering
\begin{tikzpicture}
\begin{axis}[
    title={MSE of the 4 best models on synthetic data},
    xlabel={Time step},
    ylabel={MSE},
    xmin=1, xmax=6,
    ymin=0, ymax=0.045,
    xtick={1,2,3,4,5,6},
    ytick={0,0.01,0.02,0.03,0.04},
    legend pos=north west,
    ymajorgrids=true,
    grid style=dashed,
]
\addplot[color=blue,]
    coordinates {(1,0.010185)(2,0.018218)(3,0.024418)(4,0.028407)(5,0.030895)(6,0.032522)};
\addplot[color=red,]
    coordinates {(1,0.002008)(2,0.002651)(3,0.003165)(4,0.003807)(5,0.004563)(6,0.005566)};
\addplot[color=green,]
    coordinates {(1,0.006149)(2,0.005259)(3,0.005530)(4,0.006509)(5,0.008535)(6,0.011473)};
\addplot[color=cyan,]
    coordinates {(1,0.003480)(2,0.004153)(3,0.004909)(4,0.005667)(5,0.006513)(6,0.007501)};
\addplot[color=black,]
    coordinates {(1,0.002606)(2,0.003169)(3,0.003559)(4,0.004013)(5,0.004703)(6,0.005736)};

\addplot[color=red,dashed,]
    coordinates {(1,0.002841)(2,0.003890)(3,0.004668)(4,0.005291)(5,0.005898)(6,0.006970)};
\addplot[color=green,dashed,]
    coordinates {(1,0.009925)(2,0.007713)(3,0.008425)(4,0.009430)(5,0.011771)(6,0.014217)};
\addplot[color=cyan,dashed,]
    coordinates {(1,0.003995)(2,0.004767)(3,0.005785)(4,0.006870)(5,0.008078)(6,0.009497)};
\addplot[color=black,dashed,]
    coordinates {(1,0.003255)(2,0.004904)(3,0.005503)(4,0.006619)(5,0.007492)(6,0.008555)};
    
\legend{Persistence,U-Net,CNN,LSTM,RNN}
\end{axis}
\end{tikzpicture}
\caption{Average MSE of each of the 4 best models and the persistence for each time step of the prediction. The dashed lines represent the MSE computed on binarized outputs.}
\label{fig:modelSynthBench}
\end{figure}
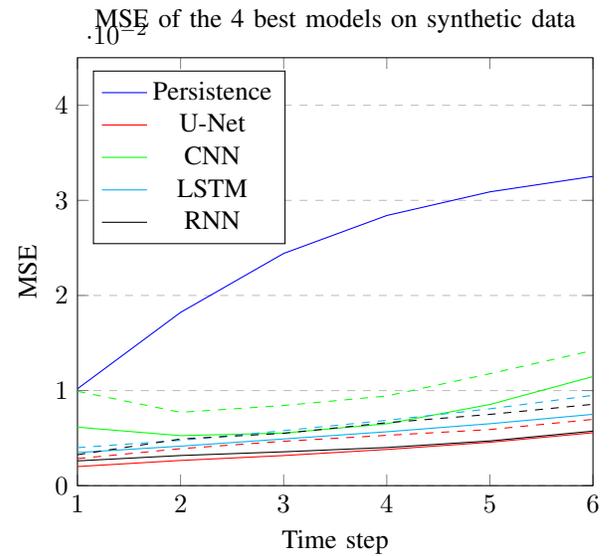

Each of the 20 models was run 3 times on each of the 6 datasets.  Figure \ref{fig:modelSynthBench} shows the average mean squared error (MSE) for the best model of each type and for each time step. The MSE over the binarized outputs is also represented with dashed lines. We can see that each model performs better than the persistence and that the prediction gets worse as the prediction time step is further in time, which was to be expected. Our benchmark shows that the original U-Net model performed better than the other models, with a MSE at 17 percent of the persistence, and a MSE over binarized outputs at 21 percent of the persistence.

\begin{figure}[t]
  \centering
  \includegraphics[width=0.5\textwidth]{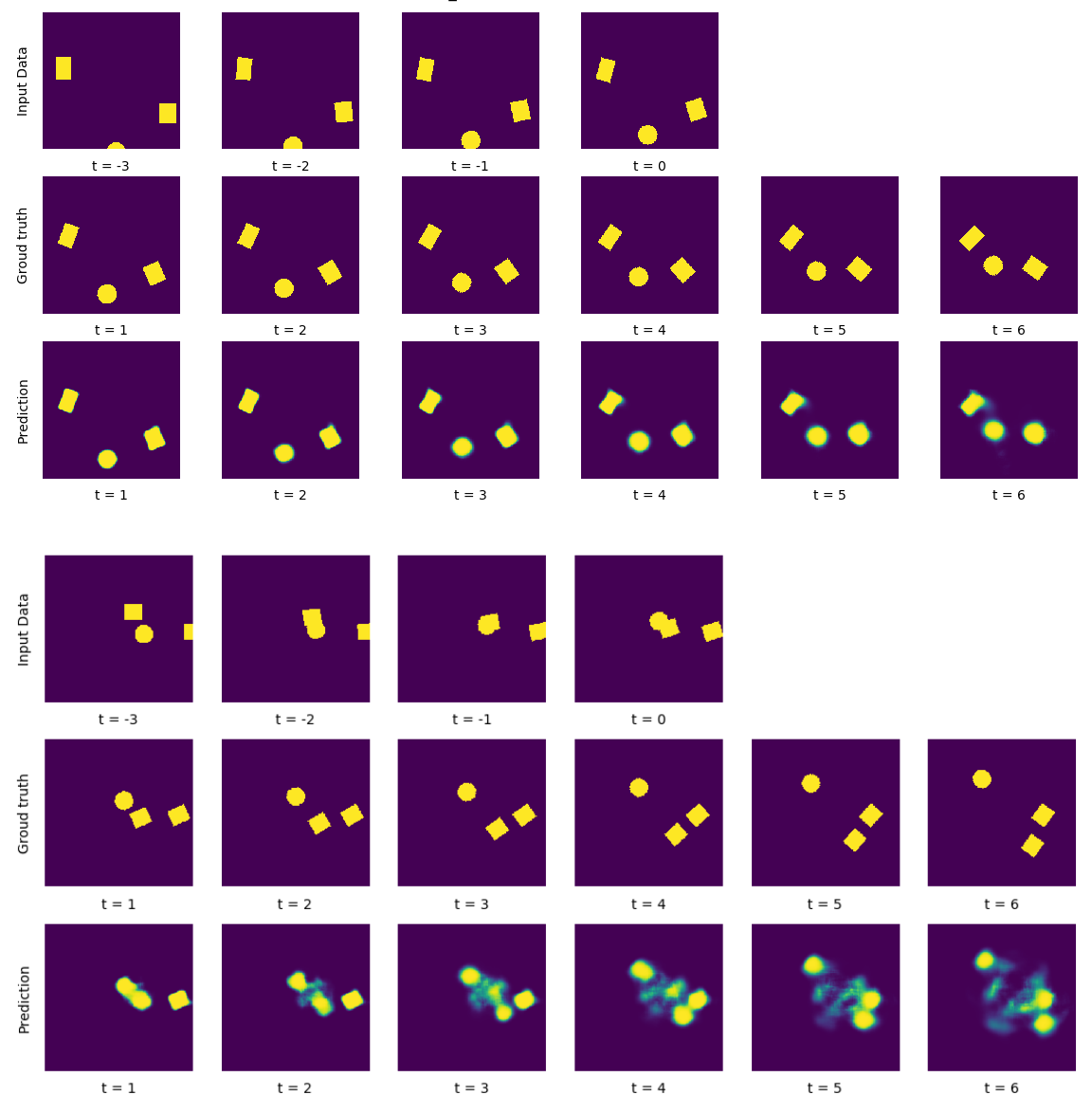}
  \caption{Two examples of U-Net's predictions : the first line is the input data, the second line is the ground truth and the third line is the output of the model.}
  \label{fig:synthRes1}
\end{figure}

When examining the outputs of the models, this first study shows promising results : Figure \ref{fig:synthRes1} shows predictions made by the U-Net model, using input from the test set. The network manages to make good predictions of the future position of the object and its shape. It even does a fairly good job when objects overlap in the input images, making them difficult to distinguish.

This study on synthetic data also reveals that the main source of error of the predictions is the architecture of the neural networks. Indeed, the exact future position of the shapes can be theoretically computed from the last two input images, at least when the shapes don't overlap. The error cannot be attributed to a weak signal in the inputs.

\section{Benchmark on real satellite data} \label{satdata}

\begin{figure}[t]
\centering
\resizebox{!}{0.8\columnwidth}{
\begin{tikzpicture}
\begin{axis}[
    title={MSE for the 4 best models on satellite data},
    xlabel={Time step (min)},
    ylabel={MSE},
    xmin=15, xmax=90,
    ymin=-0, ymax=0.3,
    xtick={15,30,45,60,75,90},
    ytick={0,0.05,0.1,0.15,0.2,0.25},
    legend pos=north west,
    ymajorgrids=true,
    grid style=dashed,
]
\addplot[color=blue,]
    coordinates {(15,0.097228)(30,0.125017)(45,0.144007)(60,0.158969)(75,0.172048)(90,0.183035)};
    
\addplot[color=red,]
    coordinates {(15,0.053228)(30,0.066889)(45,0.076967)(60,0.084975)(75,0.092092)(90,0.098159)};
\addplot[color=green,]
    coordinates {(15,0.060833)(30,0.072903)(45,0.081940)(60,0.090815)(75,0.098411)(90,0.104633)};
\addplot[color=cyan,]
    coordinates {(15,0.063139)(30,0.073429)(45,0.083607)(60,0.092693)(75,0.100649)(90,0.107282)};
\addplot[color=black,]
    coordinates {(15,0.061075)(30,0.074556)(45,0.084235)(60,0.096763)(75,0.107243)(90,0.113285)};
    
\addplot[color=red,dashed,]
    coordinates {(15,0.074002)(30,0.093589)(45,0.108022)(60,0.119442)(75,0.129743)(90,0.138677)};
\addplot[color=green,dashed,]
    coordinates {(15,0.082896)(30,0.099692)(45,0.113601)(60,0.125846)(75,0.136501)(90,0.145644)};
\addplot[color=cyan,dashed,]
    coordinates {(15,0.087416)(30,0.102051)(45,0.116400)(60,0.128994)(75,0.140283)(90,0.149806)};
    \addplot[color=black,dashed,]
    coordinates {(15,0.081668)(30,0.099434)(45,0.113222)(60,0.127803)(75,0.140324)(90,0.149110)};
    
\legend{Persistence,U-Net,CNN,LSTM,RNN}
\end{axis}
\end{tikzpicture}}
\caption{Average MSE of each model and the persistence for each time step of the prediction. The dashed lines represent the MSE computed on binarized values.}
\label{fig:mseBench}
\end{figure}
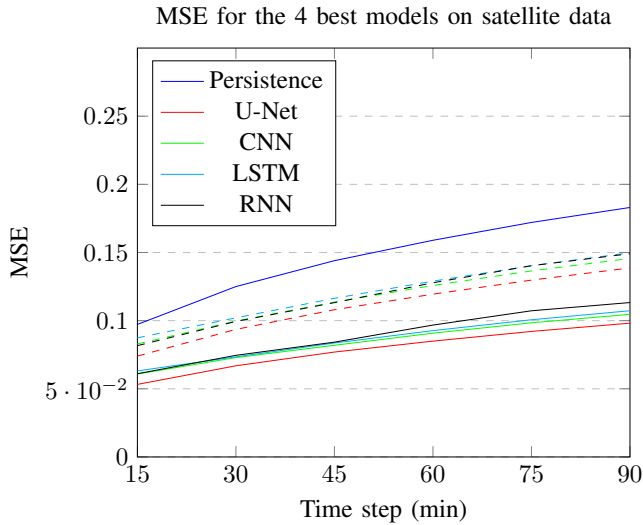

\begin{figure}[t]
  \centering
  \includegraphics[width=0.5\textwidth]{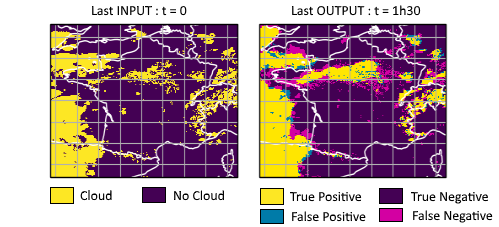}
  \caption{Example of prediction made by the U-Net architecture. Left figure: last input image received by the model. Right figure: last output of the model.}
  \label{fig:predSat}
\end{figure}

Based on the first study on synthetic data, we selected four models, the best of each category, and tested them on the satellite data described in section \ref{methodology}.

Figure \ref{fig:mseBench} shows the average MSE at each time step for each model, as well as the MSE computed over binarized images, compared to the persistence. We can see that each model performed better than persistence, even on the first prediction, and that the U-Net got an error 53 percent smaller than persistence. In addition, the MSE on binarized outputs is 75 percent smaller than persistence. All models managed significant improvement of persistence and were able to learn to predict the movement of cloud covers. Figure \ref{fig:predSat} shows an example of prediction made by the U-Net model, comparing the last input of the model and its last prediction.

We can see that stationary clouds (East and South-West) are correctly predicted. In addition, the global movement of clouds is well represented : the North-West clouds are correctly moving to the North. Finally, the cloud appearing in the North is well approximated despite its blurry edges.

\section{Temporal depth of inputs} \label{input1}

Then, we studied the influence of the number of input images on the performance of the model. For this experiment, we only studied the U-Net model, as it performed better than the others on the previous experiments. The idea was to check if the model could use older input data to make better predictions, or if the last four images considered so far were enough to predict the future positions of clouds.

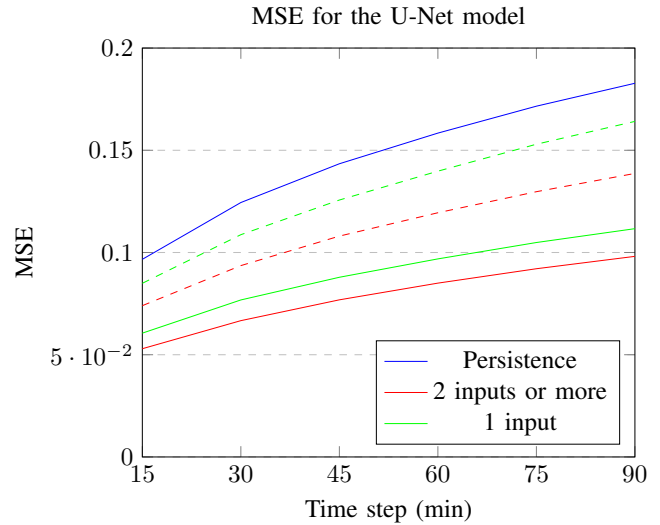
\begin{figure}[t]
\centering
\resizebox{!}{0.8\columnwidth}{
\begin{tikzpicture}
\begin{axis}[
    title={MSE for the U-Net model},
    xlabel={Time step (min)},
    ylabel={MSE},
    xmin=15, xmax=90,
    ymin=0, ymax=0.2,
    xtick={15,30,45,60,75,90},
    ytick={0,0.05,0.1,0.15,0.2},
    legend pos=south east,
    ymajorgrids=true,
    grid style=dashed,
    cycle list name=black white,
]
\addplot[color=blue,]
    coordinates {(15,0.096665)(30,0.124439)(45,0.143403)(60,0.158399)(75,0.171568)(90,0.182770)};
\addplot[color=red,]
    coordinates {(15,0.052944)(30,0.066669)(45,0.076848)(60,0.085021)(75,0.092062)(90,0.098113)};
\addplot[color=green,]
    coordinates {(15,0.060563)(30,0.076794)(45,0.087887)(60,0.096874)(75,0.104883)(90,0.111657 	)};
\addplot[color=red,dashed,]
    coordinates {(15,0.074002)(30,0.093589)(45,0.108022)(60,0.119442)(75,0.129743)(90,0.138677)};
\addplot[color=green, dashed,]
    coordinates {(15,0.084937)(30,0.108712)(45,0.125663)(60,0.139782)(75,0.152955)(90,0.164128)};
\legend{Persistence,2 inputs or more,1 input}
\end{axis}
\end{tikzpicture}}
\caption{Average MSE of the U-Net model and the persistence, with different number of input images, for each time step of the prediction.}
\label{fig:mse_inputs}
\end{figure}

\begin{figure}[t]
  \centering
  \includegraphics[width=0.5\textwidth]{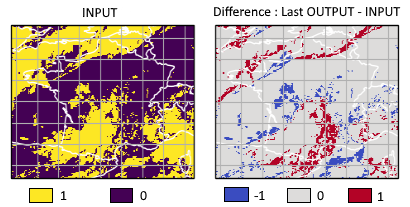}
  \caption{Single input experiment: the left image represents the last input given to the model; the right image is the difference between the last output ($t$+1h30) and the input. Appearing clouds are represented in red and disappearing clouds are in blue.}
  \label{fig:1input}
\end{figure}

To this end, we varied the length of input data by changing the number of input images given to the model, which ranged between 1 and 6, corresponding to 15 minutes to 1h30 of temporal depth. This experiment showed that using more than 2 input images did not offer any benefits compared to using only 2 inputs: the variations in the MSE and in the output images were not significant. This result supports the conclusions of \cite{b10} on rainfall nowcasting which shows that reducing the input's temporal context from 90 minutes to 30 minutes does not affect the performances significantly.

In addition, using only one input image still results in a better performance than persistence in terms of MSE, even when the model shouldn't have any indication of cloud movements. Figure \ref{fig:mse_inputs} represents the evolution of MSE regarding the number of input images considered and figure \ref{fig:1input} shows an example of output from the U-Net trained with only one input image.  We can see that the model added clouds on the east of existing clouds (in red), which tends to make them move to the east. As the model was only trained on France area, our hypothesis is that the model learned a part of the local climatology of France where clouds are mostly moving from west to east. Therefore, the model can predict the most probable movement of cloud masses even with only one input image.

\section{Size and shape evolution} \label{cloudevolution}

\begin{figure}[b]
  \centering
  \includegraphics[width=0.5\textwidth]{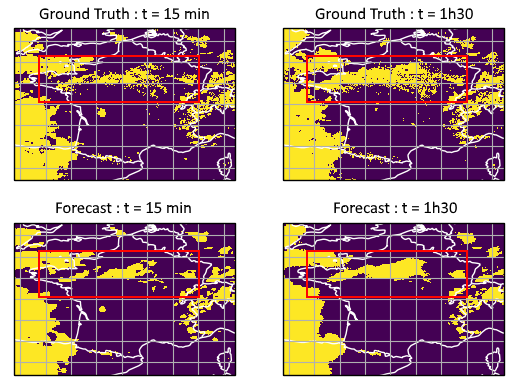}
  \caption{Ground truth: in the area enclosed by a red line, we observe a cloud increasing in volume between +15min and +90min. Forecast: We exhibit U-net predictions at the corresponding times. The model correctly enlarges the surface of the cloud.}
  \label{fig:cloud_appearance}
\end{figure}

A well known weakness of data extrapolation techniques is
their difficulty to model size and shape evolution of clouds.
Extrapolation techniques shift existing clouds, tending to preserve initial dimensions. The ability of a model to take into account this attribute of clouds is very hard to measure accurately but simply observing the image outputted, we can get an idea. As shown in Figure \ref{fig:cloud_appearance}, we extracted multiple sequences in which deep learning models increase or decrease the size of an existing cloud. However, new clouds are not created by our architecture (which would have been surprising).

\section{Comparison with meteorological models} \label{comparison}

Finally, we compared the performance of the U-Net network and persistence with two other meteorological models :
\begin{itemize}
    \item \textbf{EXIM} : Eumetsat image extrapolation tool \cite{b8} which uses high resolution atmospheric wind vectors to forecast the position and type of clouds up to 2h30 over the whole Meteosat disk, with a time step of 15 minutes.
    \item \textbf{AROME} : METEO FRANCE non-hydrostatic very-high-resolution model \cite{b15} \cite{b16}, based on physical modelling, which forecasts several physical parameters up to 48 hours over the French territory, with a time step of 1 hour.
\end{itemize}

It should be noted that EXIM's forecasts are available 20 minutes after the beginning of the run, and AROME's are available after an hour. Thus, the first time steps of the forecasts are unusable. Meanwhile, the inference of one sample with the U-Net model can be computed in about 20 seconds on an average laptop, and persistence's forecast are available instantly, therefore all of their forecasts can be used in practice.

For this comparison, as we want to compare the performances of the models at equivalent runs, the forecasts are considered available immediately after the beginning of the run, which is favorable to EXIM and AROME.

In addition, the domain of this comparison is still the French territory but the time span is February to May 2019, as EXIM's data was not available on a larger span at the time of the study.

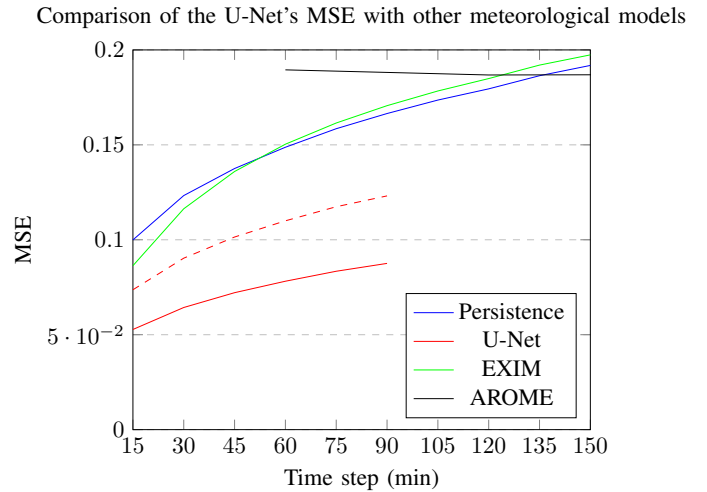
\begin{figure}[b]
\centering
\resizebox{!}{0.75\columnwidth}{
\begin{tikzpicture}
\begin{axis}[
    title={Comparison of the U-Net's MSE with other meteorological models},
    xlabel={Time step (min)},
    ylabel={MSE},
    xmin=15, xmax=150,
    ymin=0, ymax=0.2,
    xtick={15,30,45,60,75,90, 105, 120, 135, 150},
    ytick={0,0.05,0.1,0.15,0.2},
    legend pos=south east,
    ymajorgrids=true,
    grid style=dashed,
    cycle list name=black white,
]
\addplot[color=blue,]
    coordinates {(15,0.099922)(30,0.1232852)(45,0.1375326)(60,0.1487361)(75,0.1585011)(90,0.1664955)(105,0.173589)(120,0.1794931)(135,0.1864095)(150,0.1918288)};
\addplot[color=red,]
    coordinates {(15,0.05275941)(30,0.06432997)(45,0.0721333)(60,0.07816989)(75,0.0834236)(90,0.087558668)};
\addplot[color=green,]
    coordinates {(15,0.08634558)(30,0.1163287)(45,0.1359627)(60,0.1503409)(75,0.1615061)(90,0.1706357)(105,0.1783751)(120,0.1849083)(135,0.1920073)(150,0.197387)};
\addplot[color=black,]
    coordinates {(60,0.1894961)(120,0.1867991)(180,0.1870061)};
\addplot[color=red,dashed,]
    coordinates {(15,0.07367889)(30,0.09031589)(45,0.1014509)(60,0.1099887)(75,0.1174001)(90,0.1231143)};
\legend{Persistence,U-Net,EXIM,AROME}
\end{axis}
\end{tikzpicture}}
\caption{Average MSE of the U-Net model, persistence, EXIM and AROME for each time step of the forecast. The dashed line represent the MSE computed on binarized
values.}
\label{fig:compMeteo}
\end{figure}

Figure \ref{fig:compMeteo} represents the evolution of the MSE for each of the 4 models. The MSE over the binarized outputs is also represented
with a dashed line. Our comparison shows that the U-Net
model performed better than the other models. In addition, we can notice that persistence surpasses EXIM after 45 minutes, which shows the limitations of classical extrapolation methods. As expected, AROME's MSE is worse than the other models for the first time steps, but it's MSE is slowly decreasing and becomes better than EXIM and the persistence after 2 hours. Indeed, as we could expect from a physical model, AROME shows its strength on mid to long term forecasts, when extrapolation methods are no longer relevant.

\section{Conclusion and Future Work}

We provided results of multiple deep learning architectures on cloud cover nowcasting, more precisely on the data extrapolation task every 15 minutes and up to an hour and a half. All architectures proved to outperform significantly persistence at each step, in terms of MSE. So far, U-net architecture has proven to be the best for this task and we didn’t find any custom architecture to surpass it. While persistence is a baseline of interest in meteorology, U-Net also surpasses AROME, METEO FRANCE non-hydrostatic very-high-resolution model, and EXIM, Eumetsat image extrapolation tool, while demanding significantly less computation time.

As of today, our neural network architectures are not capable of predicting the emergence of new clouds. This limitation is shared by all pure data extrapolation techniques to the best of our knowledge. Adding other data sources in the hope of overcoming this limitation will be the subject of another study. We also plan to use user-tailored metrics for satellite or solar panels applications.

Finally, it will be of interest to try extending the length of the forecasts and measure when AROME's physical modelling surpasses deep learning techniques, in the hope that deep learning could eventually perform the final bridge between very short-term and mid-term forecasting.

\section*{Acknowledgment}

We would like to thank V. Chabot, M. Claudon, C. Jauffret, G. Larvor and M. Sorel from METEO FRANCE for their help and advice. We also thank EUMETSAT for the production and availability of satellite images and the \textit{Centre de M\'et\'eorologie Spatiale} for the Cloud Type product. Finally, we are deeply grateful to the \textit{Fonds de Transformation de l'Action Publique} (French government program for public services modernization) whom funding allowed the genesis of the Artificial Intelligence Laboratory of METEO FRANCE.

\vspace{12pt}

\end{document}